\documentclass[11pt]{article}

\usepackage[preprint]{acl}

\usepackage{times}
\usepackage{latexsym}

\usepackage[T1]{fontenc}

\usepackage[utf8]{inputenc}

\usepackage{microtype}

\usepackage{inconsolata}

\usepackage{graphicx}

\usepackage[dvipsnames]{xcolor}
\usepackage{times}
\usepackage{latexsym}
\usepackage{graphicx}
\usepackage{textcomp}
\usepackage{colortbl}
\usepackage{xcolor}
\usepackage{amsmath} 
\usepackage{booktabs} 
\usepackage{array}
\usepackage{graphicx} 
\usepackage{multirow}
\usepackage{color}
\usepackage{makecell} 
\usepackage{float}
\usepackage{subcaption}
\usepackage{colortbl}
\usepackage{placeins}
\usepackage[ruled,vlined]{algorithm2e}
\usepackage{comment}
\usepackage{pifont}
\usepackage{url}

\usepackage{fontawesome5}

\usepackage[T1]{fontenc}
\usepackage[utf8]{inputenc}
\usepackage{microtype}
\usepackage{inconsolata}
\usepackage{graphicx}

\usepackage{diagbox}
\usepackage{xcolor}
\usepackage{listings}
\usepackage{tcolorbox}

\title{Filter, Then Reweight: Rethinking Optimization Granularity in \\ On-Policy Distillation}

\author{\textbf{Yuying Li}$^{1*\diamond}$ \ 
\textbf{Leqi Zheng}$^{1*}$ \ 
\textbf{Yongzi Yu}$^{2}$ \ 
\textbf{Wenrui Zhou}$^{2}$  \\
\textbf{Xuchang Zhong}$^{3}$ \ 
\textbf{Xing Hu}$^{4}$ \
\textbf{Jing Jin}$^{1}$ \
\textbf{Hangjie Yuan}$^{5\dagger}$ \
\textbf{Tao Feng}$^{1\dagger}$\\
\textsuperscript{1}THU,
\textsuperscript{2}HKUST,
\textsuperscript{3}BIT,
\textsuperscript{4}Meituan,
\textsuperscript{5}ZJU\\
\tt\small liyuying25@mails.tsinghua.edu.cn fengtao.hi@gmail.com\\
\footnotesize{$^{*}$ Equal Contribution \quad $^{\dagger}$ Corresponding Author}\\
}

\begin{document}
\maketitle

\makeatletter
\def\ps@firstpage{%
  \def\@oddfoot{%
    \rlap{%
      \vbox to 0pt{%
        \vskip -22pt
        \hrule width 0.3\textwidth
        \vskip 4pt
        \hbox{\footnotesize $\diamond$\ Work done during an internship at Meituan.}%
        \vss
      }%
    }%
    \hfil\thepage\hfil
  }%
  \def\@evenfoot{\@oddfoot}%
}
\makeatother
\thispagestyle{firstpage}

\newcommand{\fireicon}{\includegraphics[height=1em]{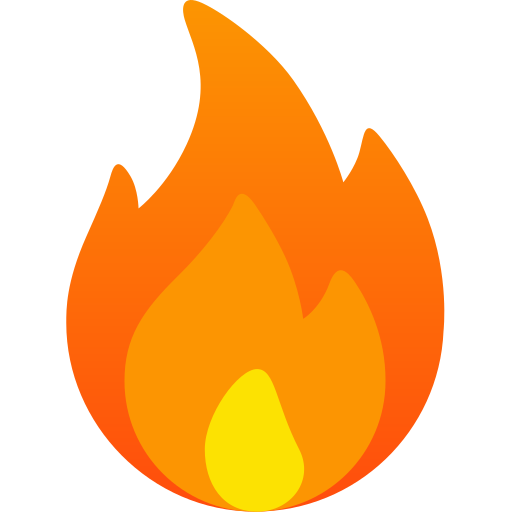}}

\begin{abstract}
On-Policy distillation (OPD) in large language models is shifting from full-trace KL supervision toward more selective training paradigms. Recent OPD methods increasingly focus on selecting which trajectories to learn from, which tokens are most informative, and which supervision signals are most reliable. Motivated by this trend, we rethink optimization granularity of OPD and propose \fireicon\ FiRe-OPD (Filter, then Reweight), which jointly adjusts supervision signals at both trajectory and token levels. In details, FiRe-OPD first filters trajectories to remove low-quality rollout samples, and then applies soft reweighting within the retained trajectories to emphasize informative tokens. Compared with hard token selection, FiRe-OPD leverages a soft-weighting mechanism to effectively mitigate information loss and enhance optimization stability, thereby achieving finer-grained OPD optimization. We validate the effectiveness of FiRe-OPD across strong-to-weak, single-teacher, and multi-teacher settings, and demonstrate its superiority over recent token-level OPD methods ( (e.g., +6.25 on AIME 2024 in strong-to-weak, +18.81 on Miner in multi-teacher). Our code is available at https://github.com/YuYingLi0/FiRe-OPD.
\end{abstract}

\begin{figure*}[htbp]
    \centering
    \includegraphics[width=1.0\linewidth]{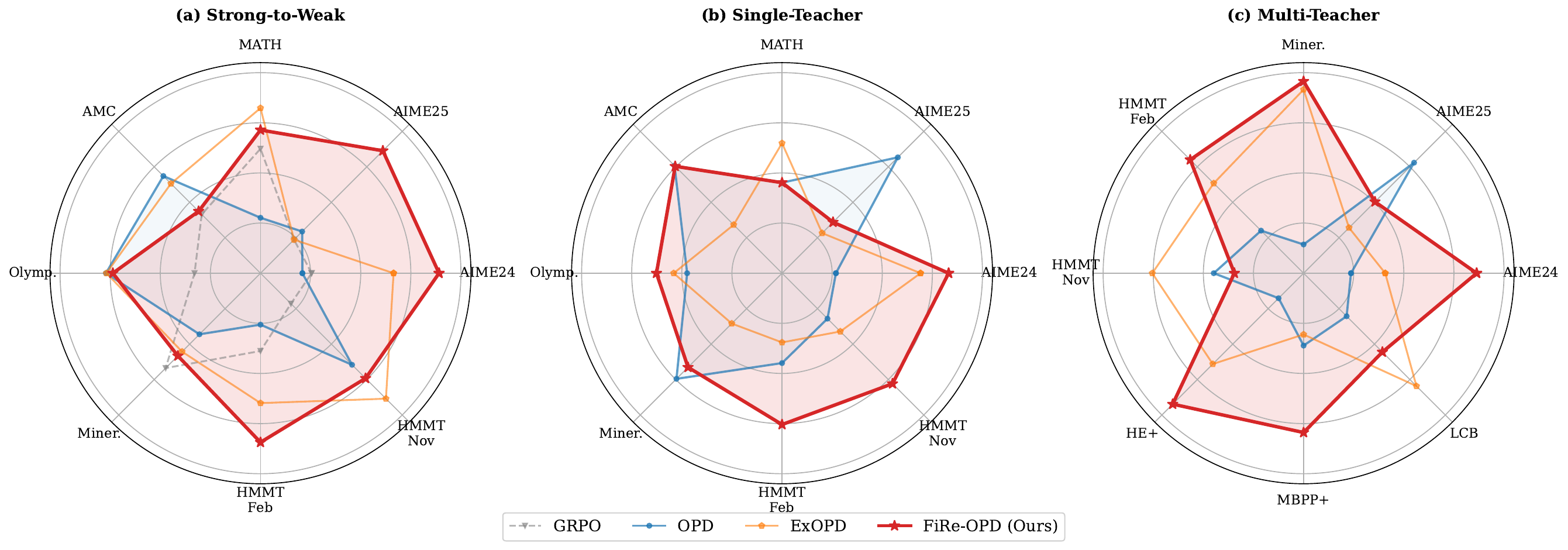}
    \caption{Performance comparison across three distillation scenarios. \textbf{FiRe-OPD} (\textcolor{red}{red}) achieves the most balanced and expansive coverage across all benchmarks}
    \label{fig: radar}
    \vspace{-2mm}
\end{figure*}

\section{Introduction}

On-policy distillation (OPD) has emerged as a compelling post-training paradigm for transferring reasoning capabilities from teacher models to smaller student models.
Unlike supervised fine-tuning, OPD avoids the train-inference distribution mismatch by learning on student-generated trajectories, while providing denser token-level supervision than reinforcement learning's sparse outcome rewards~\cite{hpd, opcd, reopd,wu2026lightning, fu2026revisiting, scope, veto, opdsurvey}.
These advantages have made OPD a widely adopted approach in reasoning-intensive tasks.

However, standard OPD applies uniform full-trajectory KL supervision, which has inherent limitations in both optimization granularity and signal reliability. Not all trajectories and tokens carry equal learning value, and critical rollouts and informative tokens should be assigned greater importance during optimization. Recognizing this, selective optimization granularity distillation has become a growing trend in recent OPD research.

EOPD~\cite{eopd} identifies that high teacher entropy causes unstable learning signals and switches to forward KL at high-entropy token positions. TIP~\cite{tip} selects tokens based on student entropy and teacher-student divergence through hard filtering rules. ExOPD~\cite{exopd} reinterprets OPD as KL-constrained RL and introduces a global reward scaling factor. Uni-OPD~\cite{uniopd} addresses unreliable supervision through outcome-guided margin calibration at the trajectory level. But existing works suffer from two key limitations:

\begin{table}[t]
\centering
\small
\renewcommand{\arraystretch}{1.1}
\caption{Overview of OPD methods across granularities and techniques, and the scope of \fireicon\ \textbf{FiRe-OPD}.}
\setlength{\tabcolsep}{3pt}
\begin{tabular}{lcc|ccc}
\toprule
\multirow{2}{*}{\textbf{Method}} & \multicolumn{2}{c|}{\textbf{Granularity}} & \multicolumn{3}{c}{\textbf{Technique}} \\
\cmidrule(lr){2-3} \cmidrule(l){4-6}
 & \textbf{Traj.} & \textbf{Tok.} & \textbf{T-Conf.} & \textbf{S-Conf.} & \textbf{Soft-W.} \\
\midrule
OPD & \textcolor{red}{\ding{55}} & \textcolor{red}{\ding{55}} & \textcolor{red}{\ding{55}} & \textcolor{red}{\ding{55}} & \textcolor{red}{\ding{55}} \\
EOPD & \textcolor{red}{\ding{55}} & \textcolor{green!60!black}{\ding{51}} & \textcolor{green!60!black}{\ding{51}} & \textcolor{red}{\ding{55}} & \textcolor{red}{\ding{55}} \\
TIP & \textcolor{red}{\ding{55}} & \textcolor{green!60!black}{\ding{51}} & \textcolor{red}{\ding{55}} & \textcolor{green!60!black}{\ding{51}} & \textcolor{red}{\ding{55}} \\
ExOPD & \textcolor{red}{\ding{55}} & \textcolor{red}{\ding{55}} & \textcolor{red}{\ding{55}} & \textcolor{red}{\ding{55}} & \textcolor{red}{\ding{55}} \\
REOPOLD & \textcolor{red}{\ding{55}} & \textcolor{green!60!black}{\ding{51}} & \textcolor{green!60!black}{\ding{51}} & \textcolor{red}{\ding{55}} & \textcolor{red}{\ding{55}} \\
\midrule
\rowcolor{gray!15}
\textbf{\fireicon\ FiRe-OPD} & \textcolor{green!60!black}{\ding{51}} & \textcolor{green!60!black}{\ding{51}} & \textcolor{green!60!black}{\ding{51}} & \textcolor{green!60!black}{\ding{51}} & \textcolor{green!60!black}{\ding{51}} \\
\bottomrule
\end{tabular}
\label{tab:method_comparison}
\end{table}

\textbf{Limitation 1.} \textit{Granularity isolation.} Existing methods operate at either the trajectory or token level, focusing on a single dimension of signal quality (e.g., teacher confidence or student state), without jointly modeling both granularities or exploiting their complementary in OPD.

\textbf{Limitation 2.} 
\textit{Hard selection strategies.} Most token-level methods rely on hard selection to remove tokens during OPD, which induces non-smooth optimization and permanently discards potentially useful supervision signals, thereby weakening learning robustness. Table~\ref{tab:method_comparison} provides a systematic comparison of existing OPD methods along these dimensions.

In this work, we propose \fireicon\ \textbf{FiRe-OPD} (\textbf{Fi}lter, then \textbf{Re}weight), a unified framework that performs trajectory-level filtering and token-level importance weighting from a dual perspective of teacher confidence and student confusion. At the trajectory level, FiRe-OPD filters out rollouts where the teacher assigns low overall likelihood, indicating a large teacher-student distribution gap where the teacher's supervision is unreliable. At the token level, FiRe-OPD assigns continuous importance weights by multiplicatively combining teacher confidence and student confusion, concentrating learning on positions where the teacher provides reliable guidance and the student has genuine need. This soft weighting preserves gradient contributions from all positions proportional to their informativeness, enabling fine-grained, adaptive supervision that accounts for both \textit{what the teacher can teach} and \textit{what the student needs to learn}.

In summary, our contributions are 3-fold:

(i) We propose FiRe-OPD, a unified framework that jointly performs trajectory-level filtering and token-level soft reweighting, enabling fine-grained and selective OPD.

(ii) We show that optimization granularity is critical in OPD: hard filtering is more effective at the trajectory level, whereas soft token weighting surpasses hard token selection at the token level.

(iii) We show that the superiority of FiRe-OPD across strong-to-weak, single-teacher, and multi-teacher distillation settings on various benchmark.

\begin{figure*}[htbp]
    \centering
    \includegraphics[width=1.0\linewidth]{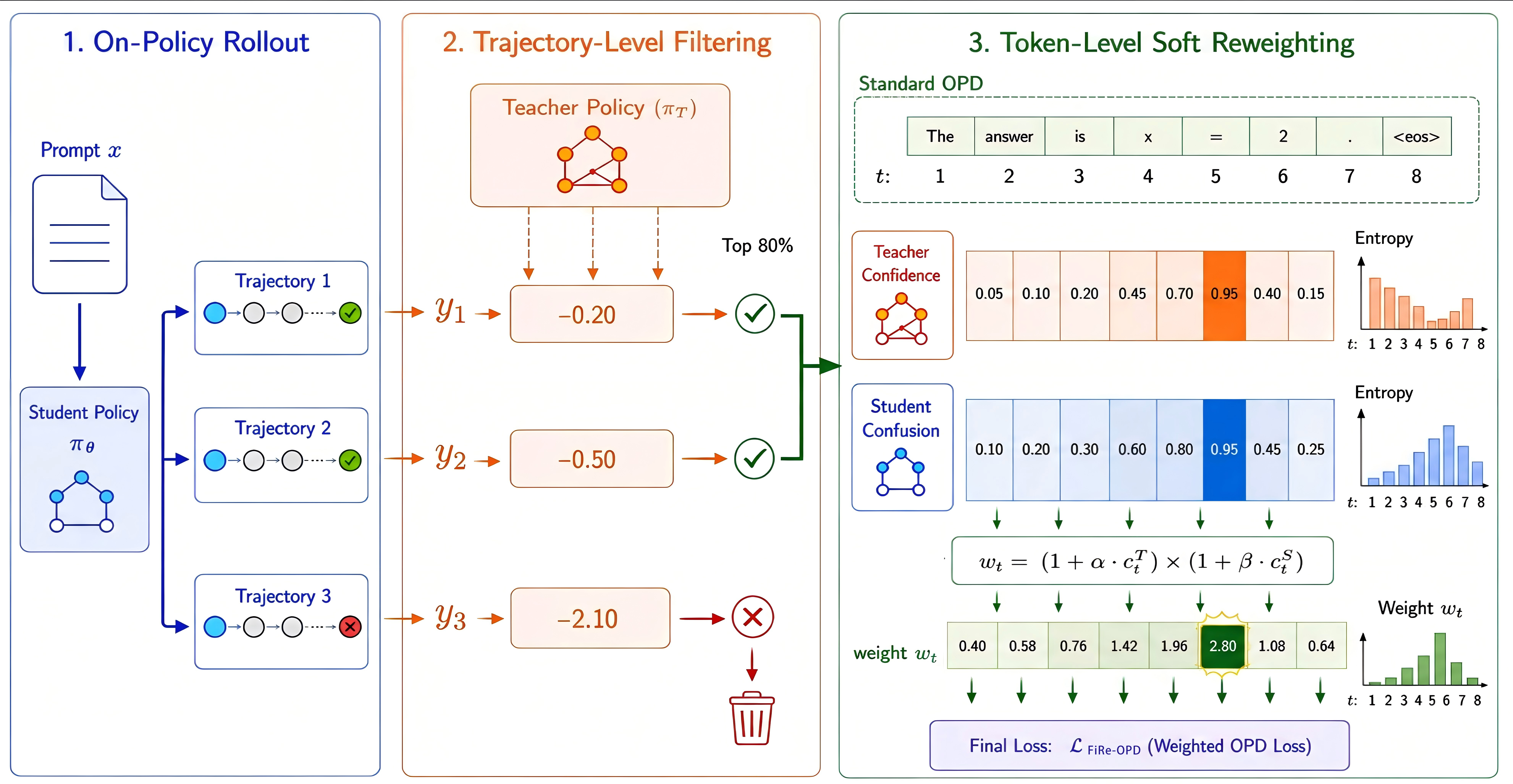}
    \caption{Overview of \fireicon\ \textbf{FiRe-OPD} that performs trajectory-level filtering and token-level importance weighting.}
    \label{fig: main}
    \vspace{-2mm}
\end{figure*}

\section{Related Work}
\textbf{Off-policy Distillation.} Knowledge distillation (KD) transfers knowledge from a stronger teacher to a smaller student model. Classical KD trains the student to match the teacher's output distribution, while sequence-level KD uses complete teacher-generated responses as supervision~\cite{hinton2015distilling, kim2016sequence}. In the LLM era, KD has evolved toward broader capability transfer like reasoning and alignment.~\cite{gu2024minillm, ko2025distillm, he2025kd, liu2024ddk}. However, most off-policy KD methods rely on teacher-generated trajectories, leading to exposure bias. These limitations motivate OPD, which directly supervises the student under its own generation distribution.

\noindent \textbf{On-Policy Distillation.} OPD has recently emerged as an effective paradigm for post-training. Prior studies show that reverse-KL-style objectives and supervision on student-generated mistakes can improve open-ended generation and reasoning tasks~\cite{gu2024minillm, agarwal2024policy}. Recent work further studies how to make OPD scalable, stable, and generalizable through reward extrapolation, entropy-aware objectives, reasoning-prefix acceleration, competence-aware curricula, divergence constraints, and rollout mixture distillation
~\cite{exopd,eopd,zhang2026fast, luo2026demystifying,uniopd}. Meanwhile, OPD has also been extended to self-distillation~\cite{opsd, xu2026paced, skillopsd, opsddoes, opsdl, sdft}, hybrid RL-distillation frameworks\cite{luffy,sdpo, rlad, ding2026hdpo, rlsd, opft}, multimodal distillation\cite{videoopd, speechopd,piopd, vold}, agentic settings\cite{tcod}, and embodied learning\cite{vlaopd}. 
Recent token-selection methods attempt to reduce noisy supervision by discarding low-value tokens, but hard selection may lose useful information and produce brittle optimization signals. Our work addresses this limitation through adaptive trajectory and token-level weighting, filtering low-quality trajectories and softly modulating token-level distillation intensity.

\section{Methodology}

\subsection{Preliminaries}

\begin{table*}[htbp]
\centering
\small
\renewcommand{\arraystretch}{1.1}
\caption{Strong-to-weak distillation results (Avg@8). Best results among OPD methods are in \textbf{bold}. \textcolor{red}{Red}/\textcolor{green!50!black}{green} denotes improvement/decline vs. OPD.}
\begin{tabular}{lcccccccc|c}
\toprule
\rowcolor{blue!10}
\textbf{Method} & \textbf{AIME24} & \textbf{AIME25} & \textbf{MATH} & \textbf{AMC} & \textbf{Olymp.} & \textbf{Miner.} & \textbf{\makecell{HMMT\\Feb}} & \textbf{\makecell{HMMT\\Nov}} & \textbf{Avg} \\
\midrule
\multicolumn{10}{c}{\textit{Strong-to-Weak: Qwen3-30B-A3B-Instruct $\rightarrow$ Qwen3-4B}} \\
\midrule
Student (Base) & 21.67 & 22.50 & 83.65 & 67.19 & 51.80 & 39.48 & 12.50 & 7.08 & 38.23 \\
Teacher & 76.67 & 63.33 & 97.22 & 95.94 & 78.32 & 47.47 & 45.00 & 60.00 & 70.49 \\
\midrule
+ SFT & 25.42 & 22.92 & 85.82 & 70.31 & 54.60 & 40.81 & 13.75 & 12.92 & 40.82 \\
+ GRPO & 55.00 & 48.33 & 93.20 & 93.06 & 68.69 & 43.73 & 29.17 & 35.42 & 58.33 \\
\midrule
+ OPD & 54.58 & 48.75 & 91.25 & \textbf{93.92} & 70.62 & 43.01 & 28.33 & 39.17 & 58.70 \\
+ ExOPD & 58.75 & 48.33 & \textbf{94.35} & 93.75 & 70.61 & 43.38 & 30.83 & \textbf{41.25} & 60.16 \\
+ TIP & 59.58 & 49.58 & 92.19 & 93.60 & 70.66 & \textbf{43.70} & 29.58 & 40.00 & 59.86 \\
+ REOPOLD & 57.50 & 46.67 & 93.95 & 92.19 & 70.16 & 43.20 & 29.17 & \textbf{41.25} & 59.26 \\
+ EOPD & 52.92 & 49.17 & 93.40 & 92.81 & \textbf{70.92} & 42.97 & 27.08 & 39.17 & 58.56 \\
\midrule
\rowcolor{orange!15}
\textbf{+ \fireicon\ FiRe-OPD (Ours)} & \textbf{60.83} & \textbf{52.92} & 93.73 & 93.13 & 70.47 & 43.47 & \textbf{32.08} & 40.00 & \textbf{60.83} \\
\rowcolor{orange!5}
$\Delta$ vs OPD & \textcolor{red}{+6.25} & \textcolor{red}{+4.17} & \textcolor{red}{+2.48} & \textcolor{green!50!black}{-0.79} & \textcolor{green!50!black}{-0.15} & \textcolor{red}{+0.46} & \textcolor{red}{+3.75} & \textcolor{red}{+0.83} & \textcolor{red}{+2.13} \\
\bottomrule
\end{tabular}

\label{tab:strong_to_weak}
\end{table*}

We first introduce the standard on-policy distillation (OPD) framework. Let $\pi_\theta$ denote the student model and $\pi_T$ denote the teacher model. At each training iteration, the student generates rollouts from its current policy given a set of prompts $\{x_i\}$:
\begin{equation}
    y \sim \pi_\theta(\cdot | x)
\label{eq:rollout}
\end{equation}

The teacher then provides token-level supervision on these student-generated trajectories. Standard OPD formulates this as a policy optimization problem using PPO-style clipped objectives, where the token-level advantage is defined as the teacher-student log-likelihood ratio:
\begin{equation}
    a_t = \log \pi_T(y_t | x, y_{<t}) - \log \pi_{\theta_{\text{old}}}(y_t | x, y_{<t})
\label{eq:advantage}
\end{equation}

This advantage encourages the student to increase probability on tokens that the teacher assigns higher likelihood than the student's old policy. The policy gradient loss is:
\begin{equation}
    \mathcal{L}_{\text{OPD}}
    =
    -\frac{1}{T}
    \sum_{t=1}^{T}
    \min\left(
        r_t a_t,\;
        \text{clip}(r_t, 1-\epsilon, 1+\epsilon) a_t
    \right).
\label{eq:opd_loss}
\end{equation}
where $r_t = \frac{\pi_\theta(y_t | x, y_{<t})}{\pi_{\theta_{\text{old}}}(y_t | x, y_{<t})}$ is the importance sampling ratio and the $clip$ constrains $r_t$ to $[1-\epsilon, 1+\epsilon]$ to prevent excessively large policy updates. We set $\epsilon = 0.2$. Standard OPD applies this objective uniformly across all trajectories and all token positions, treating every supervision signal equally.

\subsection{\fireicon\ FiRe-OPD}

Standard OPD applies uniform supervision across all trajectories and token positions, which is suboptimal because distillation signal quality varies significantly at both levels. 
As illustrated in Figure~\ref{fig: main}, FiRe-OPD addresses this through two complementary mechanisms: trajectory-level filtering and token-level soft reweighting.

\textbf{Proposition 1.} \textit{What signal best reflects the importance of a trajectory?} 

Some works use outcome correctness~\cite{scope, uniopd} or reward scores to select trajectories. However, these approaches require external verifiers and do not directly reflect the teacher's supervision capability on a given path.

We observe that the teacher's log-probability on a student-generated trajectory reflects the distributional alignment between teacher and student on that path. A low teacher log-probability indicates a large distribution gap---the student's reasoning path diverges significantly from what the teacher would produce. In such cases, regardless of whether the trajectory is objectively correct, the teacher's token-level guidance along this path is unreliable: the teacher is effectively being asked to supervise a reasoning style it is unfamiliar with. Forcing distillation on these high-divergence trajectories can introduce noisy or even contradictory gradients, leading to negative transfer rather than effective learning.

Based on this insight, we define the \textbf{trajectory-level importance score} as the teacher's normalized log-probability over a rollout $y = (y_1, \ldots, y_T)$ given prompt $x$:
\begin{equation}
    s(y) = \frac{1}{T} \sum_{t=1}^{T} \log \pi^*(y_t | x, y_{<t})
\label{eq:traj_score}
\end{equation}
We rank all rollouts within a training batch by $s(y)$ and discard the bottom $p\%$ (we use $p=20$ by default). Only the surviving trajectories proceed to the token-level optimization stage. This filtering ensures that distillation occurs only on trajectories where the teacher can provide coherent supervision---paths that lie within the teacher's competence region, where its token-level signals are most likely to be meaningful and consistent.

\begin{table*}[htbp]
\centering
\small
\renewcommand{\arraystretch}{1.1}
\caption{Single-teacher distillation results (Avg@8). Best results among OPD methods are in \textbf{bold}. \textcolor{red}{Red}/\textcolor{green!50!black}{green} denotes improvement/decline vs. OPD.}
\resizebox{\linewidth}{!}{
\begin{tabular}{lcccccccc|c}
\toprule
\rowcolor{blue!10}
\textbf{Method} & \textbf{AIME24} & \textbf{AIME25} & \textbf{MATH} & \textbf{AMC} & \textbf{Olymp.} & \textbf{Miner.} & \textbf{\makecell{HMMT\\Feb}} & \textbf{\makecell{HMMT\\Nov}} & \textbf{Avg} \\
\midrule
\multicolumn{10}{c}{\textit{Single-Teacher: Qwen3-4B-Non-Thinking-RL-Math $\rightarrow$ Qwen3-4B}} \\
\midrule
Student (Base) & 21.67 & 22.50 & 83.65 & 67.19 & 51.80 & 39.48 & 12.50 & 7.08 & 38.23 \\
Teacher & 56.25 & 46.67 & 93.40 & 91.56 & 68.06 & 47.52 & 30.83 & 35.83 & 58.77\\
\midrule
+ OPD & 57.92 & \textbf{57.50} & 94.69 & \textbf{95.17} & 47.66 & \textbf{68.81} & 32.50 & 35.42 & 61.21 \\
+ ExOPD & 60.42 & 54.58 & \textbf{95.25} & 93.44 & 47.84 & 67.28 & 32.08 & 35.83 & 60.84 \\
\midrule
\rowcolor{orange!15}
\textbf{+ \fireicon\ FiRe-OPD (Ours)} & \textbf{61.25} & 55.00 & 94.69 & 95.15 & \textbf{48.07} & 68.49 & \textbf{33.75} & \textbf{37.50} & \textbf{61.74} \\
\rowcolor{orange!5}
$\Delta$ vs OPD & \textcolor{red}{+3.33} & \textcolor{green!50!black}{-2.50} & \textcolor{red}{+0.00} & \textcolor{green!50!black}{-0.02} & \textcolor{red}{+0.41} & \textcolor{green!50!black}{-0.32} & \textcolor{red}{+1.25} & \textcolor{red}{+2.08} & \textcolor{red}{+0.53} \\
\bottomrule
\end{tabular}
}
\label{tab:single_teacher}
\end{table*}

\begin{table*}[htbp]
\centering
\small
\renewcommand{\arraystretch}{1.1}
\caption{Multi-teacher distillation results (Qwen3-4B-Non-Thinking-RL-Math + Qwen3-4B-Non-Thinking-RL-Code $\rightarrow$ Qwen3-4B-Non-Thinking). Best results are in \textbf{bold}.}
\resizebox{\linewidth}{!}{
\begin{tabular}{lcccccc cccc}
\toprule
& \multicolumn{6}{c}{\textbf{Math Reasoning (Avg@8)}} & \multicolumn{4}{c}{\textbf{Code Generation (pass@1)}} \\
\cmidrule(lr){2-7} \cmidrule(lr){8-11}
\rowcolor{blue!10}
\multirow{-2}{*}{\textbf{Method}} & \textbf{AIME24} & \textbf{AIME25} & \textbf{Miner.} & \textbf{\makecell{HMMT\\Feb}} & \textbf{\makecell{HMMT\\Nov}} & \textbf{Avg} & \textbf{HE+} & \textbf{MBPP+} & \textbf{LCB} & \textbf{Avg} \\
\midrule
Student (Base) & 21.67 & 22.50 & 39.48 & 12.50 & 7.08 & 20.65 & 79.90 & 64.60 & 17.57 & 54.02 \\
Teacher & 76.67 & 63.33 & 47.47 & 45.00 & 60.00 & 58.49 & 79.90 & 72.00 & 27.85 & 59.92 \\
\midrule
OPD & 59.58 & 57.08 & 48.53 & 32.50 & 37.50 & 47.04 & 82.93 & 69.58 & 26.86 & 59.79 \\
+ ExOPD & 60.83 & 55.00 & 66.39 & 34.17 & \textbf{38.75} & 51.03 & 89.00 & 69.31 & \textbf{29.28} & 62.53 \\
\midrule
\rowcolor{orange!15}
\textbf{+ \fireicon\ FiRe-OPD} & \textbf{64.17} & \textbf{55.83} & \textbf{67.34} & \textbf{35.00} & 37.08 & \textbf{51.88} & \textbf{92.70} & \textbf{71.69} & 28.10 & \textbf{64.16} \\
\rowcolor{orange!5}
$\Delta$ vs OPD & \textcolor{red}{+4.59} & \textcolor{green!50!black}{-1.25} & \textcolor{red}{+18.81} & \textcolor{red}{+2.50} & \textcolor{green!50!black}{-0.42} & \textcolor{red}{+4.84} & \textcolor{red}{+9.77} & \textcolor{red}{+2.11} & \textcolor{red}{+1.24} & \textcolor{red}{+4.37} \\
\bottomrule
\end{tabular}
}
\label{tab:multi_teacher}
\end{table*}

\textbf{Proposition 2.} \textit{ What makes a token position informative for distillation?} 

Existing works either focus on a single signal---teacher entropy alone~\cite{reopold} or student entropy alone~\cite{tip}---or apply hard truncation that discards tokens entirely below a fixed threshold. These approaches either miss one important dimension of signal quality or irreversibly lose gradient information from positions that still carry partial learning value. In contrast, we argue that a position is most informative for distillation when two conditions are jointly satisfied: the teacher is confident (providing reliable guidance) and the student is confused (indicating genuine learning need). This motivates a unified, soft weighting scheme that integrates both signals simultaneously.

Based on this, we define the \textbf{token-level importance weight} using two complementary signals. Teacher confidence $c_t^T$ measures how reliable the teacher's guidance is at position $t$:
\begin{equation}
    c_t^T = 1 - \frac{H(\pi^*(\cdot | x, y_{<t}))}{\max_{t' \in \mathcal{B}} H(\pi^*(\cdot | x, y_{<t'}))}
\label{eq:teacher_conf}
\end{equation}
where $H(\cdot)$ denotes the entropy of the output distribution and $\max_{t' \in \mathcal{B}}$ is the empirical maximum over all valid token positions in the current batch $\mathcal{B}$. Student confusion $c_t^S$ measures how much the student needs guidance at position $t$:
\begin{equation}
    c_t^S = \frac{H(\pi_\theta(\cdot | x, y_{<t}))}{\max_{t' \in \mathcal{B}} H(\pi_\theta(\cdot | x, y_{<t'}))}
\label{eq:student_conf}
\end{equation}

The token-level importance weight combines both signals multiplicatively:
\begin{equation}
    w_t = (1 + \alpha \cdot c_t^T) \times (1 + \beta \cdot c_t^S)
\label{eq:token_weight}
\end{equation}
where $\alpha, \beta \geq 0$ are hyperparameters controlling the sensitivity to each respective factor (we use $\alpha = \beta = 1.0$ by default throughout all experiments). The weighted advantage for each token is then obtained by normalizing the raw weights within each trajectory to preserve gradient scale:
\begin{equation}
    \tilde{a}_t = \frac{w_t}{\frac{1}{T}\sum_{t'=1}^{T} w_{t'}} \cdot a_t
\label{eq:weighted_adv}
\end{equation}
The normalization ensures that the total gradient magnitude remains stable across different trajectories. The final policy gradient loss is:

\begin{equation}
    \mathcal{L}_{\text{FiRe-OPD}}
    \!=\!
    -\frac{1}{T}\!\sum_{t=1}^{T}
    \min\!\bigl(
        r_t \tilde{a}_t,\,
        \mathrm{clip}(r_t,\, 1\!-\!\epsilon,\, 1\!+\!\epsilon)\tilde{a}_t
    \bigr)
\label{eq:fire_opd_loss}
\end{equation}

This design concentrates learning effort on positions where the teacher is confident yet the student remains confused, while still preserving gradient contributions from all positions in proportion to their relative informativeness.

\begin{table*}[htbp]
\centering
\small
\renewcommand{\arraystretch}{1.1}
\caption{Ablation study on component contributions (Avg@8, Strong-to-Weak setting). Best results are in \textbf{bold}.}
\resizebox{\linewidth}{!}{
\begin{tabular}{lcccccccc|c}
\toprule
\rowcolor{blue!10}
\textbf{Method} & \textbf{AIME24} & \textbf{AIME25} & \textbf{MATH} & \textbf{AMC} & \textbf{Olymp.} & \textbf{Miner.} & \textbf{\makecell{HMMT\\Feb}} & \textbf{\makecell{HMMT\\Nov}} & \textbf{Avg} \\
\midrule
OPD (Base) & 54.58 & 48.75 & 91.25 & \textbf{93.92} & 70.62 & 43.01 & 28.33 & 39.17 & 58.70 \\
\midrule
\rowcolor{orange!15}
\textbf{\fireicon\ FiRe-OPD (Full)}  & \textbf{60.83} & \textbf{52.92} & 93.73 & 93.13 & 70.47 & \textbf{43.47} & \textbf{32.08} & \textbf{40.00} & \textbf{60.83} \\
\midrule
w/o Traj. Filter & 56.92 & 49.88 & 93.54 & 91.19 & 70.41 & 43.12 & 28.33 & 38.42 & 58.99 \\
w/o Teacher Conf. & 58.33 & 52.08 & \textbf{94.08} & 90.31 & \textbf{70.64} & 42.46 & 28.75 & 37.92 & 59.32 \\
w/o Student Conf. & 59.08 & 49.87 & 93.51 & 91.34 & 70.33 & 43.00 & 31.17 & 37.92 & 59.53 \\
Traj. Filter Only & 54.58 & 51.67 & 93.85 & 91.88 & 69.96 & 42.88 & 30.42 & 39.17 & 59.30 \\
\bottomrule
\end{tabular}
}
\label{tab:ablation_component}
\end{table*}

\begin{figure*}[htbp]
    \centering
    \includegraphics[width=1.0\linewidth]{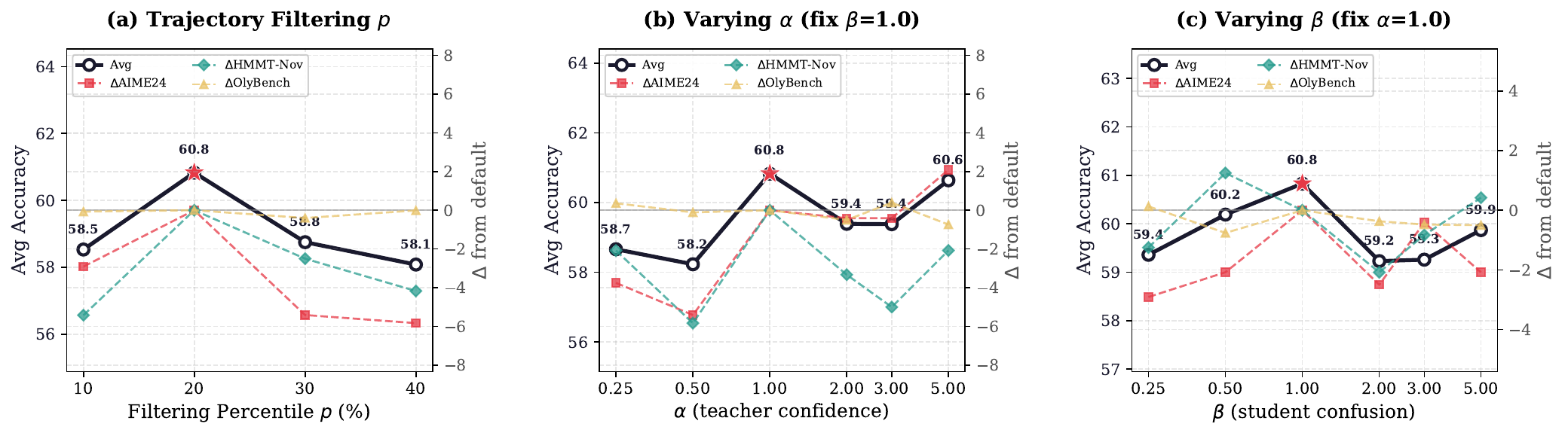}
    \caption{Hyperparameter sensitivity analysis. The solid black line (left axis) shows Avg accuracy across all benchmarks; dashed colored lines (right axis) show per-benchmark deviations ($\Delta$) from the default setting. (a)~Trajectory filtering percentile $p$ exhibits a clear peak at $p{=}20\%$. (b)~Performance is robust for $\alpha \geq 1.0$ but degrades notably at small values. (c)~$\beta$ shows minimal sensitivity across the full range, confirming that student confusion weighting is robust to its scaling.}
    \label{fig:param_sen}
    \vspace{-2mm}
\end{figure*}

\section{Experiment}

\subsection{Experimental Setup}

\paragraph{Models.} To demonstrate the generalizability of FiRe-OPD, we evaluate across three distillation scenarios: (i) \textit{Strong-to-Weak}, where Qwen3-30B-A3B-Instruct serves as the teacher and Qwen3-4B-Non-Thinking~\cite{yang2025qwen3} as the student, testing the ability to bridge large capacity gaps; (ii) \textit{Single-Teacher}, where Qwen3-4B-Non-Thinking-RL-Math~\cite{exopd} teaches Qwen3-4B-Non-Thinking, testing transfer efficiency between same models; and (iii) \textit{Multi-Teacher}, where Qwen3-4B-Non-Thinking-RL-Math and Qwen3-4B-Non-Thinking-RL-Code~\cite{exopd} jointly supervise Qwen3-4B, testing the ability to integrate heterogeneous domain expertise.

\paragraph{Training Data.} For the strong-to-weak and single-teacher scenarios, we use the DeepMath-103K~\cite{he2025deepmath} dataset filtered to difficulty level 6, following~\cite{exopd}. For the multi-teacher scenario, we use the multi-teacher training dataset from~\cite{exopd}, which combines mathematical and code-domain data.

\paragraph{Training Details.} We train for 3 epochs (165 steps total) with a batch size of 1024, learning rate of $1 \times 10^{-6}$, and maximum response length of 16384. During rollout, we sample with temperature 1.0 and top-$p$ = 1.0. For FiRe-OPD-specific hyperparameters, we set $\alpha = \beta = 1.0$ and trajectory filtering percentile $p = 20\%$. Training is conducted on 8$\times$A100 80GB GPUs.

\paragraph{Evaluation.} For mathematical reasoning, we evaluate on eight benchmarks spanning a range of difficulty levels: AIME 2024, AIME 2025, MATH-500~\cite{hendrycks2021measuring}, AMC 2023, OlympiadBench~\cite{he2024olympiadbench}, MinervaMATH~\cite{lewkowycz2022solving}, HMMT 2025 Feb, and HMMT 2025 Nov~\cite{hmmt}. We sample 8 responses per problem with temperature 1.0 and report Avg@8 accuracy. For code generation, we evaluate on three widely-used benchmarks: HumanEval+~\cite{humanevalmbpp}, MBPP+~\cite{humanevalmbpp}, and LiveCodeBench (v6 only, February 2025--May 2025)~\cite{livecodebench}, and report pass@1 accuracy.

\paragraph{Baselines.} We compare against standard OPD and five recent improvements: ExOPD~\cite{exopd}, TIP~\cite{tip}, REOPOLD~\cite{reopold}, EOPD~\cite{eopd}, and Uni-OPD~\cite{uniopd}. ExOPD uses the official open-source implementation; TIP, REOPOLD, and EOPD are reproduced by us. All methods are trained under the same data, model, and compute budget for fair comparison. We also report SFT and GRPO results as reference.

\begin{table*}[htbp]
\centering
\small
\renewcommand{\arraystretch}{1.1}
\caption{Ablation on soft weighting vs. hard truncation (Avg@8, Strong-to-Weak setting). ``Hard/Soft'' denotes trajectory-level filtering and token-level weighting strategy respectively. Best results are in \textbf{bold}.}
\resizebox{\linewidth}{!}{
\begin{tabular}{lcccccccc|c}
\toprule
\rowcolor{blue!10}
\textbf{Method} & \textbf{AIME24} & \textbf{AIME25} & \textbf{MATH} & \textbf{AMC} & \textbf{Olymp.} & \textbf{Miner.} & \textbf{\makecell{HMMT\\Feb}} & \textbf{\makecell{HMMT\\Nov}} & \textbf{Avg} \\
\midrule
\rowcolor{orange!15}
\textbf{\fireicon\ FiRe-OPD (Hard + Soft)} & \textbf{60.83} & \textbf{52.92} & 93.73 & \textbf{93.13} & 70.47 & 43.47 & \textbf{32.08} & \textbf{40.00} & \textbf{60.83} \\
\midrule
Hard + Hard & 57.92 & 46.25 & \textbf{93.95} & 90.62 & 68.53 & \textbf{43.15} & 31.67 & 33.75 & 58.23 \\
Soft + Soft & 57.92 & 48.75 & 93.75 & 90.94 & \textbf{70.75} & \textbf{43.15} & 28.33 & 35.83 & 58.68 \\
Soft + Hard & 55.42 & 51.25 & 93.35 & 89.38 & 68.86 & 43.01 & 30.42 & 36.67 & 58.55 \\
\bottomrule
\end{tabular}
}
\label{tab:ablation_soft_hard}
\end{table*}

\subsection{Main Results}

\paragraph{Strong-to-Weak Distillation.} Table~\ref{tab:strong_to_weak} presents results for distilling from a 30B teacher to a 4B student. FiRe-OPD achieves the highest average accuracy of 60.83\%, outperforming the strongest baseline ExOPD (60.16\%) by 0.67 points and standard OPD (58.70\%) by 2.13 points. The improvements are particularly pronounced on challenging competition-level benchmarks: +6.25 on AIME 2024, +4.17 on AIME 2025, +3.75 on HMMT Feb, and +2.48 on MATH-500. We also observe that FiRe-OPD substantially outperforms both SFT (which barely improves over the base model) and GRPO, confirming the advantage of dense teacher supervision combined with adaptive weighting. Compared to other token-level methods, FiRe-OPD consistently outperforms TIP (+0.97 avg), REOPOLD (+1.57 avg), and EOPD (+2.27 avg), demonstrating the effectiveness of our method.

\paragraph{Single-Teacher Distillation.} Table~\ref{tab:single_teacher} shows results where the teacher and student share the same architecture size, representing a minimal distribution gap scenario. FiRe-OPD achieves the highest average accuracy of 61.74\%, improving over standard OPD (61.21\%) by 0.53 points and ExOPD (60.84\%) by 0.90 points, with notable gains on competition-level tasks such as AIME 2024 (+3.33) and HMMT Nov (+2.08). The consistent gains confirm that FiRe-OPD remains beneficial even when the teacher-student distribution gap is small.

\paragraph{Multi-Teacher Distillation.} Table~\ref{tab:multi_teacher} presents results where two domain-specialized teachers (math and code) jointly supervise one student. FiRe-OPD achieves the best math reasoning average of 51.88\% (+4.84 over OPD) and code generation average of 64.16\% (+4.37 over OPD). The gains are substantial across both domains: +18.81 on MinervaMAT and +4.59 on AIME 2024 for math, +9.77 on HumanEval+ and +2.11 on MBPP+ for code. Notably, FiRe-OPD enables the student to substantially surpass both teachers on code tasks (92.70 vs.\ 79.90 on HumanEval+), demonstrating effective knowledge integration beyond simple imitation.

\paragraph{Cross-Scenario Analysis.} In the strong-to-weak setting, gains concentrate on competition-level benchmarks, confirming that trajectory filtering effectively removes low-quality rollouts that would otherwise corrupt learning on hard problems. Single-teacher distillation yields uniform but modest improvements given the smaller capacity gap, while the multi-teacher setting exhibits the most dramatic gains (+4.84 avg on math), as filtering and adaptive weighting naturally resolves conflicts between heterogeneous teachers. Overall, FiRe-OPD scales gracefully with distillation difficulty---whether from capacity gaps, task complexity, or supervision heterogeneity.

\subsection{Ablation Studies}
\label{sec:ablation}

To gain deeper understanding of how each mechanism contributes to FiRe-OPD's effectiveness, we conduct comprehensive ablations analyzing the contribution of each component, the sensitivity to hyperparameters, and the effectiveness of soft weighting versus hard truncation. All ablations are performed in the Strong-to-Weak setting.

\paragraph{Component Ablation.} Table~\ref{tab:ablation_component} presents results when removing individual components. The full FiRe-OPD (60.83) significantly outperforms all ablated variants. Removing student confusion causes the largest drop (-2.24), followed by trajectory filtering (-1.84), while removing teacher confidence has the smallest impact (-0.96). This reveals an asymmetric role: student confusion is the dominant token-level signal determining \textit{where} the student needs help, while teacher confidence serves as a complementary quality filter. Trajectory filtering alone (59.30) already outperforms OPD (58.70), but combining it with token-level weighting yields further gains (+1.53), confirming that both granularities contribute complementarily.

\begin{figure*}[htbp]
    \centering
    \includegraphics[width=1.0\linewidth]{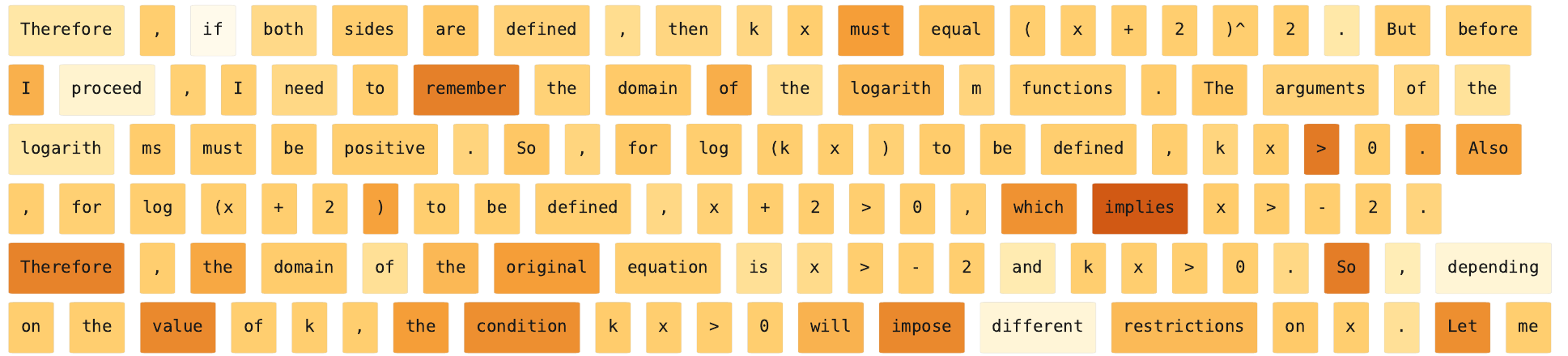}
    \caption{Case Study. Visualization of \textbf{FiRe-OPD}'s token-level weight allocation on a math reasoning trajectory}
    \label{fig: case}
    \vspace{-2mm}
\end{figure*}

\begin{figure}[htbp]
\centering
\begin{minipage}[h]{0.49\textwidth}
    \centering    
    \includegraphics[width=\linewidth]{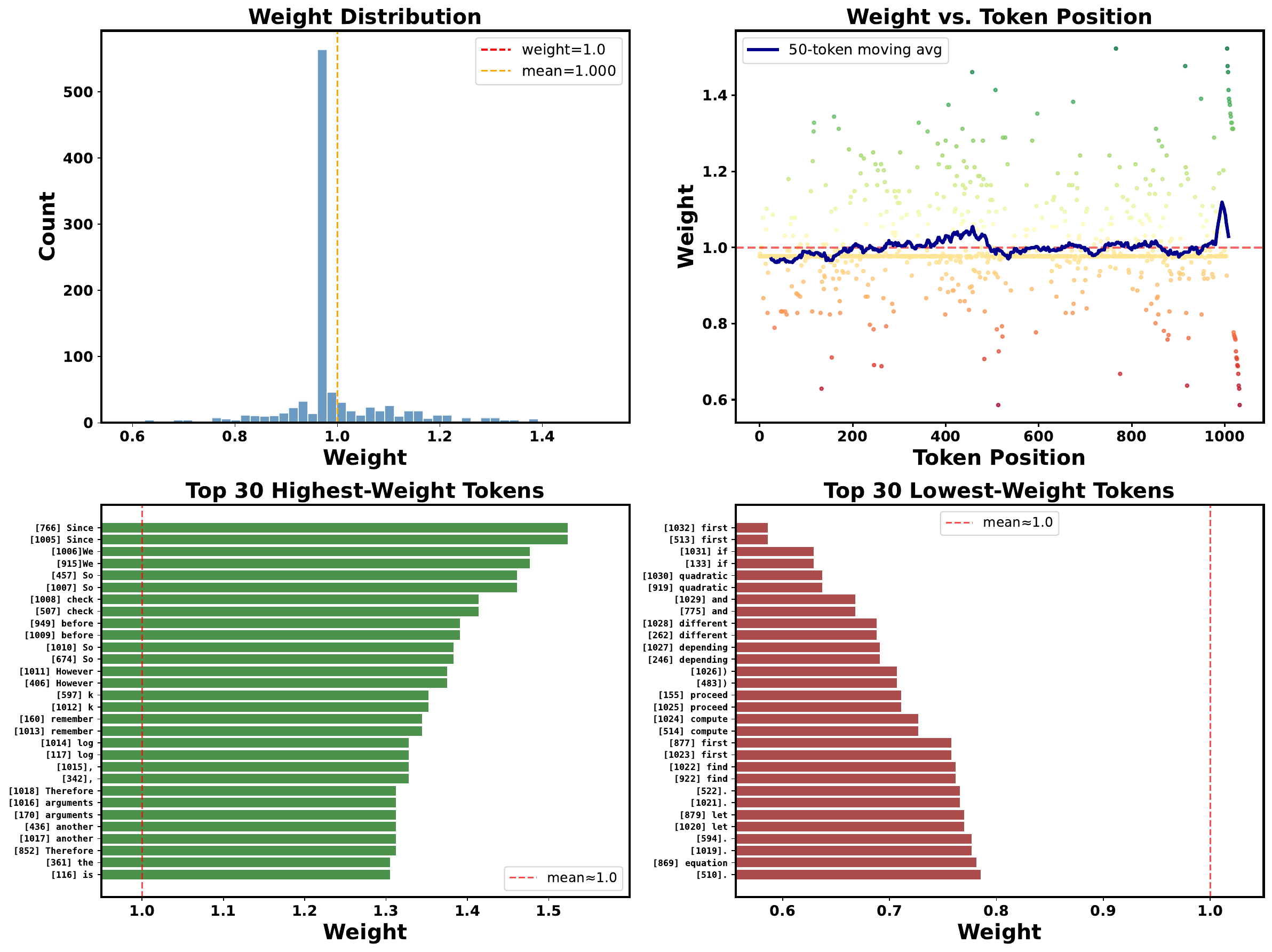}
    \captionof{figure}{Statistical Analysis of Weight Allocation.}
    \label{fig:weight}

\end{minipage}
\end{figure}

\paragraph{Hyperparameter Sensitivity.} Figure~\ref{fig:param_sen} visualizes sensitivity to the three hyperparameters. For filtering percentile $p$ (Figure~\ref{fig:param_sen}a), performance peaks at $p=20\%$, with both under-filtering ($p=10\%$: 58.53) and over-filtering ($p=40\%$: 58.08) degrading results. For $\alpha$ (Figure~\ref{fig:param_sen}b), performance is robust for $\alpha \geq 1.0$ but degrades at small values ($\alpha=0.5$: 58.23), confirming teacher confidence is necessary albeit secondary. For $\beta$ (Figure~\ref{fig:param_sen}c), performance shows minimal sensitivity across the full range, confirming that student confusion weighting is robust to its scaling. Complete per-benchmark results are provided in Tables~\ref{tab:ablation_skip} and~\ref{tab:ablation_alpha_beta_full}.

\paragraph{Soft Weighting vs. Hard Truncation.} Table~\ref{tab:ablation_soft_hard} compares four combinations of trajectory-level and token-level strategies (Hard=discrete filtering/selection, Soft=continuous weighting). FiRe-OPD's design (Hard trajectory filtering + Soft token weighting) achieves the best average of 60.83, outperforming Hard+Hard (58.23), Soft+Soft (58.68), and Soft+Hard (58.55). This validates two design choices: (1) At the trajectory level, hard filtering is superior to soft weighting, because low-quality trajectories should be completely removed rather than down-weighted, as even reduced gradients from unreliable paths can accumulate noise. (2) At the token level, soft weighting outperforms hard selection, since tokens exist on a continuum of informativeness, and preserving gradient contributions proportional to their value yields better optimization than binary keep-or-discard decisions.

\subsection{Case Study: Token Weight Visualization.}
To provide intuitive understanding of how FiRe-OPD allocates learning effort, we visualize the token-level weights on a representative mathematical reasoning trajectory in Figure~\ref{fig: case}, where darker shading indicates higher weight. The highest weights are assigned to reasoning transition tokens such as ``Therefore,'' ``implies,'' and ``So''---positions where the teacher confidently knows the next direction but the student remains uncertain. Conversely, numerical values, operators, and variable names receive minimal weights, as both models are highly confident on these tokens once the reasoning path is determined. Notably, the weighting is genuinely context-dependent: the same token ``the'' receives different weights depending on whether it introduces a critical reasoning conclusion or appears in a routine phrase.

Figure~\ref{fig:weight} further corroborates this pattern statistically through four complementary views of the learned token-level weight distribution. The upper-left panel displays the overall weight histogram, which is sharply peaked around 1.0, confirming purely redistributive reweighting that preserves total gradient magnitude. The upper-right panel presents a positional analysis showing weights increasing toward the end of the trajectory, where reasoning conclusions and final answers typically reside. The two bottom panels list representative tokens at the extremes of the weight spectrum: the highest-weight tokens are dominated by reasoning connectives (``Since,'' ``So,'' ``However,'' ``Therefore'') and metacognitive cues (``check,'' ``remember''), while the lowest-weight tokens consist of procedural words (``proceed,'' ``compute,'' ``find'') and formulaic punctuation. Together, these visualizations reveal that FiRe-OPD automatically identifies the distillation bottleneck as reasoning strategy selection---deciding \textit{what to do next}---rather than computational execution, and concentrates learning effort on decision points where teacher guidance provides the greatest informational value.

\section{Conclusion}
We propose FiRe-OPD, a dual-granularity framework for on-policy distillation that filters low-confidence trajectories and assigns continuous token-level weights based on teacher confidence and student confusion. Experiments across three distillation scenarios on math reasoning and code generation benchmarks demonstrate consistent improvements over standard OPD and recent baselines. Ablation studies reveal that teacher and student signals contribute asymmetrically across granularities, and that the two levels favor different selection strategies---hard filtering for trajectories and soft weighting for tokens.

\section{Limitations}
While FiRe-OPD demonstrates consistent improvements, the design space for adaptive distillation granularity remains largely unexplored. Our current approach treats each token independently without modeling how erroneous prefixes may degrade subsequent teacher signals---a prefix-aware weighting scheme could yield further gains. Additionally, intermediate granularities such as step-level or segment-level weighting, which align more naturally with chain-of-thought structure, represent promising directions. We leave these explorations to future work.

\bibliography{main}

\clearpage
\newpage

\appendix

\noindent
\begin{minipage}{\textwidth}
\centering
\small
\renewcommand{\arraystretch}{1.1}
\captionof{table}{Full ablation on $\alpha$ and $\beta$ (Avg@8, Strong-to-Weak setting). Default: $\alpha=1.0, \beta=1.0$.}
\label{tab:ablation_alpha_beta_full}
\resizebox{\textwidth}{!}{
\begin{tabular}{l|cccccc|cccccc}
\toprule
\rowcolor{blue!10}
& \multicolumn{6}{c|}{\textbf{Varying $\alpha$ (fix $\beta=1.0$)}} & \multicolumn{6}{c}{\textbf{Varying $\beta$ (fix $\alpha=1.0$)}} \\
\rowcolor{blue!10}
\textbf{Benchmark} & $\alpha$=0.25 & $\alpha$=0.5 & $\alpha$=1.0 & $\alpha$=2.0 & $\alpha$=3.0 & $\alpha$=5.0 & $\beta$=0.25 & $\beta$=0.5 & $\beta$=1.0 & $\beta$=2.0 & $\beta$=3.0 & $\beta$=5.0 \\
\midrule
AIME24 & 57.08 & 55.42 & \textbf{60.83} & 60.42 & 60.42 & 62.92 & 57.92 & 58.75 & \textbf{60.83} & 58.33 & 60.42 & 58.75 \\
AIME25 & 47.50 & 48.75 & 52.92 & 49.17 & 51.25 & \textbf{53.33} & 48.33 & \textbf{53.33} & 52.92 & 50.00 & 48.75 & 50.00 \\
MATH500 & 93.20 & 93.65 & 93.73 & 93.70 & 93.47 & 93.88 & 93.50 & 93.58 & 93.73 & 94.03 & 93.40 & \textbf{94.27} \\
AMC2023 & 90.62 & 92.19 & 93.13 & 90.62 & 90.62 & \textbf{93.75} & 92.19 & 92.19 & 93.13 & 91.25 & 90.31 & 90.94 \\
OlympiadBench & \textbf{70.83} & 70.36 & 70.47 & 69.97 & 70.86 & 69.73 & 70.60 & 69.71 & 70.47 & 70.10 & 69.99 & 69.97 \\
MinervaMAT & 43.34 & 42.97 & 43.47 & \textbf{44.12} & 42.97 & 42.78 & 43.57 & 43.15 & 43.47 & 42.65 & 42.88 & \textbf{44.21} \\
HMMT-Feb & 28.75 & 28.33 & \textbf{32.08} & 30.42 & 30.42 & 30.83 & 30.00 & 29.58 & \textbf{32.08} & 29.58 & 29.17 & 30.42 \\
HMMT-Nov & 37.92 & 34.17 & \textbf{40.00} & 36.67 & 35.00 & 37.92 & 38.75 & \textbf{41.25} & \textbf{40.00} & 37.92 & 39.17 & 40.42 \\
\midrule
Avg & 58.66 & 58.23 & \textbf{60.83} & 59.39 & 59.38 & 60.64 & 59.36 & 60.19 & \textbf{60.83} & 59.23 & 59.26 & 59.87 \\
\bottomrule
\end{tabular}
}
\end{minipage}

\section{Full Ablation Results}
\label{app:full_ablation}

\paragraph{Sensitivity to $\alpha$ and $\beta$.}
Table~\ref{tab:ablation_alpha_beta_full} presents the full per-benchmark results for the entropy-aware weighting hyperparameters $\alpha$ and $\beta$ in the strong-to-weak distillation setting. When varying $\alpha$ (teacher confidence scaling) with $\beta$ fixed at 1.0, performance peaks at $\alpha=1.0$ (60.83\% avg) and remains competitive at $\alpha=5.0$ (60.64\%), indicating that moderately amplifying teacher confidence signals is beneficial while the method is not overly sensitive to this parameter. When varying $\beta$ (student confusion scaling) with $\alpha$ fixed at 1.0, the optimal performance is again achieved at $\beta=1.0$, with a narrower range of competitive values---deviations in either direction lead to noticeable degradation on competition-level benchmarks (e.g., HMMT-Feb drops from 32.08\% to 29.17\% at $\beta=3.0$). This suggests that student confusion signals require more careful calibration than teacher confidence, as over-amplifying student uncertainty may cause the model to over-attend to positions where the learning signal is inherently noisy.

\paragraph{Sensitivity to Trajectory Filtering Percentile.}
Table~\ref{tab:ablation_skip} reports the effect of varying the trajectory-level filtering percentile $p$, which controls the fraction of lowest teacher-log-probability trajectories to discard. The optimal setting is $p=20\%$, achieving 60.83\% average accuracy. Lower filtering ($p=10\%$) retains too many off-distribution trajectories that introduce noisy gradients, while aggressive filtering ($p=30\%$ or $p=40\%$) discards potentially useful training signals, particularly hurting performance on the most challenging benchmarks---AIME 2024 drops from 60.83\% to 55.00\% at $p=40\%$, and HMMT-Feb drops from 32.08\% to 26.25\%. This confirms that a moderate filtering threshold strikes the best balance between removing harmful trajectories and preserving sufficient training diversity.

\begin{table}
\centering
\small
\renewcommand{\arraystretch}{1.1}
\caption{Ablation on trajectory filtering percentile $p$ (Avg@8). $p=20\%$ is our default.}
\begin{tabular}{lcccc}
\toprule
\rowcolor{blue!10}
\textbf{Benchmark} & $p$\textbf{=10} & $p$\textbf{=20} & $p$\textbf{=30} & $p$\textbf{=40} \\
\midrule
AIME24 & 57.92 & \textbf{60.83} & 55.42 & 55.00 \\
AIME25 & 49.17 & \textbf{52.92} & 50.42 & 47.92 \\
MATH500 & 93.60 & 93.73 & 93.45 & \textbf{94.05} \\
AMC2023 & 91.25 & 93.13 & 90.31 & 91.88 \\
OlympiadBench & 70.40 & \textbf{70.47} & 70.07 & 70.46 \\
MinervaMAT & 42.97 & 43.47 & 43.24 & \textbf{43.24} \\
HMMT-Feb & 28.33 & \textbf{32.08} & 29.58 & 26.25 \\
HMMT-Nov & 34.58 & \textbf{40.00} & 37.50 & 35.83 \\
\midrule
Avg & 58.53 & \textbf{60.83} & 58.75 & 58.08 \\
\bottomrule
\end{tabular}
\label{tab:ablation_skip}
\end{table}

\end{document}